\documentclass{article} 
\usepackage{iclr2021_conference,times}

\usepackage{amsmath,amsfonts,bm}

\def\eqref#1{equation~\ref{#1}}

\def\1{\bm{1}}

\DeclareMathAlphabet{\mathsfit}{\encodingdefault}{\sfdefault}{m}{sl}
\SetMathAlphabet{\mathsfit}{bold}{\encodingdefault}{\sfdefault}{bx}{n}

\usepackage{hyperref}
\usepackage{url}
\usepackage{graphicx}
\usepackage{xcolor}
\usepackage{subcaption}
\usepackage{multicol}
\usepackage{floatrow}
\newfloatcommand{capbtabbox}{table}[][\FBwidth]

\title{A Neural Network MCMC sampler that maximizes Proposal Entropy}

\author{Zengyi Li \\
Redwood Center for theoretical Neuroscience\\
Department of Physics\\
University of California Berkeley\\
Berkeley, CA 94720, USA\\
\texttt{zengyi\_li@berkeley.edu} \\
\And
Yubei Chen\\
Redwood Center for Theoretical Neuroscience \\
Berkeley AI Research \\
University of California Berkeley \\
Berkeley, CA 94720, USA\\
\texttt{yubeic@berkeley.edu} \\
\AND
Friedrich T. Sommer \\
Redwood Center for Theoretical Neuroscience\\
Helen Wills Neuroscience Institute \\
University of California Berkeley\\
Berkeley, CA 94720, USA\\
\texttt{fsommer@berkeley.edu}
}

\iclrfinalcopy 
\begin{document}

\maketitle

\begin{abstract}
Markov Chain Monte Carlo (MCMC) methods sample from unnormalized probability distributions and offer guarantees of exact sampling. However, in the continuous case, unfavorable geometry of the target distribution can greatly limit the efficiency of MCMC methods. Augmenting samplers with neural networks can potentially improve their efficiency. Previous neural network based samplers were trained with objectives that either did not explicitly encourage exploration, or used a L2 jump objective which could only be applied to well structured distributions. Thus it seems promising to instead maximize the proposal entropy for adapting the proposal to distributions of any shape. To allow direct optimization of the proposal entropy, we propose a neural network MCMC sampler that has a flexible and tractable proposal distribution. Specifically, our network architecture utilizes the gradient of the target distribution for generating proposals. Our model achieves significantly higher efficiency than previous neural network MCMC techniques in a variety of sampling tasks. Further, the sampler is applied on training of a convergent energy-based model of natural images. The adaptive sampler achieves unbiased sampling with significantly higher proposal entropy than Langevin dynamics sampler.
\end{abstract}

\section{Introduction}
Sampling from unnormalized distributions is important for many applications, including statistics, simulations of physical systems, and machine learning. However, the inefficiency of state-of-the-art sampling methods remains a main bottleneck for many challenging applications, such as protein folding \citep{noe2019boltzmanngenerator}, energy-based model training \citep{nijkamp2019anatomyMCMCEBM}, etc.

A prominent strategy for sampling is the Markov Chain Monte Carlo (MCMC) method \citep{neal1993MCMC}. In MCMC, one chooses a transition kernel that leaves the target distribution invariant and constructs a Markov Chain by applying the kernel repeatedly. The MCMC method relies only on the ergodicity assumption. Other than that it is general, if enough computation is performed, the Markov Chain generates correct samples from any target distribution, no matter how complex the distribution is. However, the performance of MCMC depends critically on how well the chosen transition kernel explores the state space of the problem. If exploration is ineffective, samples will be highly correlated and of very limited use for downstream applications. Despite some favorable theoretical argument on the effectiveness of some MCMC algorithms, practical implementation of them may still suffer from inefficiencies. 

Take, for example, the Hamiltonian Monte Carlo (HMC)\citep{neal2011HMChandbook} algorithm, a type of MCMC technique. HMC is regarded state-of-the-art for sampling in continuous spaces \cite{radivojevic2020modifiedHMC}. It uses a set of auxiliary momentum variables and generates new samples by simulating a Hamiltonian dynamics starting from the previous sample. This allows the sample to travel in state space much further than possible with other techniques, most of whom have more pronounced random walk behavior. Theoretical analysis shows that the cost of traversing an $d$-dimensional state space and generating an uncorrelated proposal is $O(d^{\frac{1}{4}})$ for HMC, which is lower than $O(d^{\frac{1}{3}})$ for Langevine Monte Carlo, and $O(d)$ for random walk. However, unfavorable geometry of a target distribution may still cause HMC to be ineffective because the Hamiltonian dynamics has to be simulated numerically. Numerical errors in the simulation are commonly corrected by a Metropolis-Hastings (MH) accept-reject step of a proposed sample. If the the target distribution has unfavorable geometric properties, for example, very different variances along different directions, the numerical integrator in HMC will have high error, leading to a very low accept probability \citep{betancourt2017geometricfundationHMC}. For simple distributions this inefficiency can be mitigated by an adaptive re-scaling matrix \citep{neal2011HMChandbook}. For analytically tractable distributions, one can also use the Riemann manifold HMC method \citep{girolami2011RMHMCandRMLD}. But in most other cases, the Hessian required in Riemann manifold HMC algorithm is often intractable or expensive to compute, preventing its application.

\begin{figure*}
    \centering
    \includegraphics[scale=0.85]{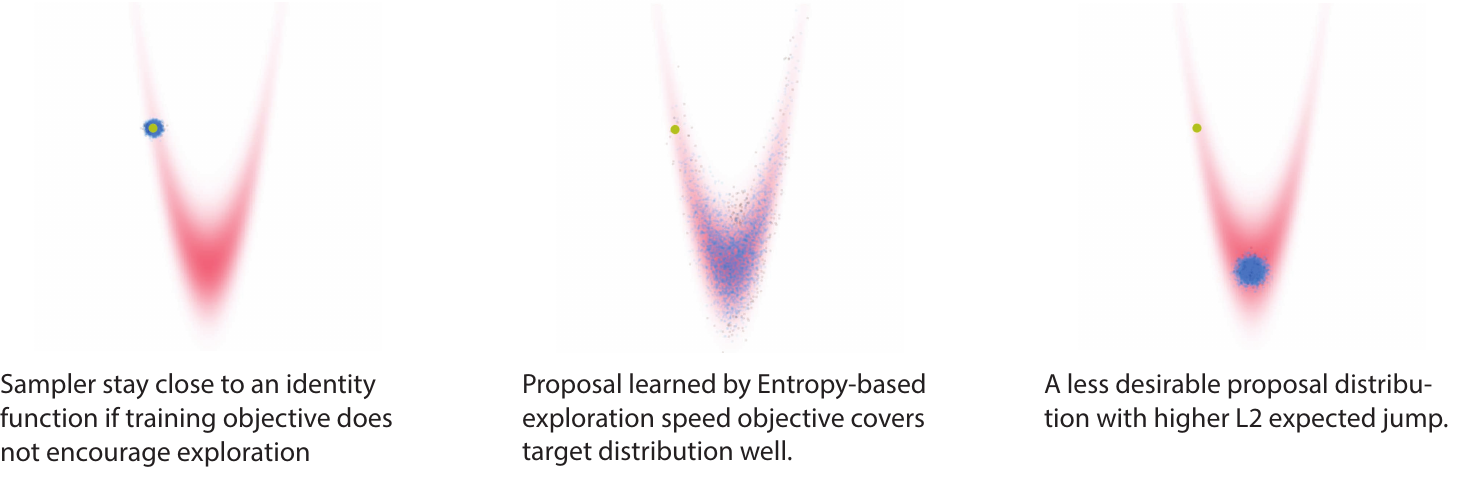}
    \caption{Illustration of learning to explore a state space. Larger yellow dot in top left is the initial point $x$, blue and black dots are accepted and rejected samples from the proposal distribution $q(x'|x)$. Solution obtained from optimizing entropy objective is close to the target distribution $p(x)$. However, we can easily construct a less desirable solution with higher L2 jump.}
    \label{fig:Fig1 illustration of L2 and entropy loss}
\end{figure*}

Recently, approaches have been proposed that possess the exact sampling property of the MCMC method, while potentially mitigating the described issues of unfavorable geometry. One approach is MCMC samplers augmented with neural networks \citep{song2017A-NICE_MC,levy2017L2HMC,gu2019NFLMC}, the  other approach is neural transport MCMC techniques \citep{hoffman2019neutra,nijkamp2020flowbackboneEBM}.
A disadvantage of these recent techniques is that their objectives optimize the quality of proposed samples, but do not explicitly encourage exploration speed of the sampler. One notable exception is L2HMC \citep{levy2017L2HMC}, a method whose objective includes the size of the expected L2 jump, thereby encouraging exploration. But the L2 expected jump objective is not very general, it only works for simple distributions (see Figure \ref{fig:Fig1 illustration of L2 and entropy loss}, and below). 

Another recent work \citep{dellaportas2019gradientbasedadaptiveMC} proposed a quite general objective to encourage exploration speed by maximizing the entropy of the proposal distribution. In continuous space, the entropy of a distribution is essentially the logarithm of its volume in state space. Thus, the entropy objective naturally encourages the proposal distribution to ``fill up" the target state space as well as possible, independent of the geometry of the target distribution. The authors demonstrated the effectiveness of this objective on samplers with simple linear adaptive parameters. 

Here we employ the entropy-based objective in a neural network MCMC sampler for optimizing exploration speed. To build the model, we design a flexible proposal distribution where the optimization of the entropy objective is tractable. Inspired by the HMC algorithm, the proposed sampler uses special architecture that utilizes the gradient of the target distribution to aid sampling. For a 2D distribution the behavior of the proposed model is illustrated in Figure \ref{fig:Fig1 illustration of L2 and entropy loss}. The sampler, trained with the entropy-based objective, generates samples that explore the target distribution quite well, while its simple to construct a proposal with higher L2 expected jump (right panel). Later we show the newly proposed method achieves significant improvement in sampling efficiency compared to previous techniques, we then apply the method to the training of an energy-based image model.

\section{Preliminary: MCMC methods, From vanilla to Learned}
\label{sec: preliminary}
Consider the problem of sampling from a target distribution $p(x) = e^{-U(x)}/Z$ defined by the energy function $U(x)$ in a continuous state space. MCMC methods solve the problem by constructing and running a Markov Chain, with transition probability $p(x'|x)$, that leaves $p(x)$ invariant. The most general invariance condition is: $p(x') = \int p(x'|x)p(x)dx$ for all $x'$, which is typically enforced by the simpler but more stringent \emph{detailed balance} condition: $p(x)p(x'|x) = p(x')p(x|x')$.

For a general distribution $p(x)$ it is difficult to directly construct $p(x'|x)$ that satisfies detailed balance, but one can easily make any transition probability satisfy it by including an additional Metropolis-Hastings accept-reject step \citep{hastings1970MHrule}. When at step $t$, we sample $x'$ from an arbitrary proposal distribution $q(x'|x^{t})$, the M-H accept-reject process accepts the new sample $x^{t+1} = x'$ with probability $A(x',x) = \min\left(1,\frac{p(x')q(x^t|x')}{p(x^t)q(x'|x^t)}\right)$. If $x'$ is rejected, the new sample is set to the previous state $x^{t+1}=x^t$. This transition kernel $p(x'|x)$ constructed from $q(x'|x)$ and $A(x',x)$ leaves any target distribution $p(x)$ invariant.

Most commonly used MCMC techniques such as Random Walk Metropolis (RWM), Metropolis-Adjusted Langevin Algorithm (MALA) and Hamiltonian Monte Carlo (HMC) use the described M-H accept-reject step to enforce detailed balance. For brevity, we will focus on MCMC methods that use the M-H step, although some alternatives exist \citep{sohl2014HMCwithoutDB}. All these methods share the requirement that the accept probability in the M-H step must be tractable to compute. For two of the mentioned MCMC methods this is indeed the case. In the Gaussian random-walk sampler, the proposal distribution is a Gaussian around the current position: $x' = x + \epsilon*\mathcal{N}(0,\mathbf{I})$, which has the form $x' = x + z$. Thus, forward and reverse proposal probability is $q(x'|x) = p_{\mathcal{N}}\left[(x'-x)/\epsilon\right]$, $q(x|x') = p_{\mathcal{N}}\left[-(x'-x)/\epsilon\right]$, Where $p_{\mathcal{N}}$ denote the density function of Gaussian. The probability ratio $\frac{q(x^t|x')}{q(x'|x^t)}$ used in the M-H step is therefore equal to $1$. In MALA the proposal distribution is a single step Langevin dynamics with step size $\epsilon$: $x'=x+z$ with $z = -\frac{\epsilon^2}{2}\partial_{x}U(x) + \epsilon N(0,\mathbf{I})$. We then have $q(x'|x) = p_{\mathcal{N}}\left[(x'-x)/\epsilon + \frac{\epsilon}{2}\partial_{x}U(x)\right]$ and $q(x|x') = p_{\mathcal{N}}\left[-(x'-x)/\epsilon + \frac{\epsilon}{2}\partial_{x'}U(x')\right]$. Both, the forward and reverse proposal probability are tractable since they are density of Gaussians evaluated at a known location.

The case for HMC is slightly more complicated. In short, HMC in its basic form also fits into the scheme of M-H sampler, although the proposal probability $q(x'|x)$ becomes intractable for general pair of $x$ and $x'$. See Section \ref{sec: related works} for a more detailed discussion.

Previous work on augmenting MCMC sampler with neural networks also relied on the M-H procedure to ensure asymptotic correctness of the sampling process, for example \citep{song2017A-NICE_MC} and \citep{levy2017L2HMC}. They used HMC style accept-reject probabilities that makes $q(x'|x)$ intractable. Here, we strive for a flexible sampler for which $q(x'|x)$ is tractable. This maintains the tractable M-H step while allowing us to train this sampler to explores the state space by directly optimizing the proposal entropy objective, which is a function of $q(x'|x)$.

\section{Gradient based sampler with tractable proposal probability}
We ``abuse'' the power of neural networks to design a sampler that is flexible and has tractable proposal probability $q(x'|x)$ between any two points. However, without the extra help of the gradient of the target distribution, the sampler will be modeling a conditional distribution $q(x'|x)$ with brute force, which is possible but requires a large model capacity. To remedy this problem, our method also use gradient of the target distribution, and we use an architecture inspired by L2HMC \citep{levy2017L2HMC} which is inspired by the HMC algorithm. The benefit of the gradient can still be somewhat subtle. To quantify the benefit we provide ablation studies of our model in the Appendix \ref{sec: ablation}. 

We restrict our sampler to the simple general form $x' = x + z$, discussed earlier. The vector $z$ is modeled by an invertible neural network: $z = f(z_{0};x,U)$, with inverse $z_{0} = f^{-1}(z;x,U)$. $f$ is an invertible function of $z$ conditioned on $x,U$ and it has tractable Jacobian determinant w.r.t. $z$. Neural networks of this type are called flow models, see \citep{kobyzev2019flowreview1,papamakarios2019flowreview2}. In our expressions, $U$ is added to emphasize the explicit dependence on the energy function. Further, $z_{0}$ is sampled from a fixed Gaussian base distribution. The proposed sampler then has tractable forward and reverse proposal probability: $q(x'|x) = p_{Z}(x'-x;x)$, $q(x|x') = p_{Z}(x-x';x')$, where $p_{Z}(z;x) = p_{\mathcal{N}}(z_{0})|\frac{\partial z}{\partial z_{0}}|^{-1}$ is the density defined by a flow model. Thus, the sampling process is straight forward. It consists first of drawing from $p_{\mathcal{N}}(z_{0})$ and then evaluating $z=f(z_{0};x,U)$ and $q(x'|x)$.  Next, the reverse $z_{0}'=f^{-1}(-z;x+z,U)$ is evaluated at $x'= x + z$ to obtain the reverse proposal probability $q(x|x')$. Finally, the sample is accepted with the standard M-H rule. Since the probabilities $q(x'|x)$ and $q(x|x')$ are conditioned on $x$ and $x'$ through a neural network, they can be highly asymmetric, allowing mixing into regions of very different probability, see for example Appendix \ref{sec: Additional results}. 

To incorporate the gradient of the target distribution into the sampler, we take inspiration from the HMC algorithm. Basic HMC  starts with drawing a random initial momentum $v^{0}$, followed by several steps of leapfrog integration. In the following, we denote the momentum variable after $n$ updates by $v^n$ and position by $x^n$. In a leapfrog step, the integrator first updates $v$ with a half step of the gradient: $v^{n\prime} = v^{n-1} - \frac{\epsilon}{2} \partial_{x} U(x^{n-1})$, followed by a full step of $x$ update: $x^{n} = x^{n-1} + \epsilon v^{n\prime}$, and another half step of $v$ update: $v^{n} = v^{n\prime} - \frac{\epsilon}{2} \partial_{x} U(x^{n})$. After several steps, the update of $x$ can be written as: $x^n = x^0 + \sum_{i=0}^{n} v^{i\prime}$, which has the form $x' = x + z$ with $z = \sum_{i}^{n} v^{i\prime} = - nv^0 - \frac{n\epsilon}{2}\left[\partial_{x} U(x^0)\right] - \epsilon\left[\sum_{i=1}^n (n-i) \partial_{x} U(x^i)\right]$. Note that, although $z$ does not have a tractable density, the equation suggests that a tractable model of $z$ that use gradient information effectively should subtract the gradient of the target distribution evaluated at intermediate point of $x$ from a sample of a base distribution $v^{0}$. 

For the invertible neural network $f$, we use an architecture similar to a non-volume preserving coupling-based flow RealNVP \citep{dinh2016RNVP}. In the coupling-based architecture, half of the components of the state vector are kept fixed and are used to update the other half through an affine transform parameterized by a neural network. We use a mask $m$ and its complement $\overline{m}$ to separate the fixed and updated half of dimensions. Further, we include the gradient of the target distribution with a negative sign in the shift term of the affine transform. We also use an element-wise scaling on the gradient term (Equation \ref{model def 1} and \ref{model def 2}). However, with a coupling based architecture, the gradient term can only depend on a subset of dimensions of the vector $z$, and it is unclear where to evaluate the gradient of the target distribution to sample effectively. Ideally, the sampler should evaluate the gradient at points far away from $x$, similar as in HMC, for traveling long distances in the state space. To achieve this effect, we use another neural network which has output $R$ and evaluate the gradient at $x+R$. In general, $R$ is allowed to depend on $x$ as well as the half of $z$ that is kept fixed. During training, $R$ learns where the gradient should be evaluated based on the available half of $z$.

To formulate our model, we define the input to network $R$ to be $\zeta_m ^n = (x,m \odot z^n)$ and the input to all other networks to be $\xi_m^n = \left(x,m \odot z^n, \partial U (x+R(\zeta_m^n))\right)$, where $\odot$ is the Hadamard product. Further, we denote the neural network outputs that parameterize the affine transform by $S(\xi_m^n)$, $Q(\xi_m^n)$ and $T(\xi_m^n)$. For notation clarity we omit dependences of the mask $m$ and all neural network terms on the step number $n$.

Additionally we introduce a scale parameter $\epsilon$, which modifies the $x$ update to $x' = x + \epsilon*z$. We also define $\epsilon' = \epsilon/(2N)$, with $N$ the total number of $z$ update steps. This parameterization makes our sampler equivalent to the MALA algorithm with step size $\epsilon$ at initialization, where the neural network outputs are zero. The resulting update rule is:
\small
\begin{align}
    \hspace{-0.1in}z^{n\prime} \hspace{-0.03in}&= m \odot z^{n-1} \hspace{-0.05in}+ \overline{m} \odot \left(z^{n-1} \hspace{-0.03in} \odot \exp[S(\xi_m^{n-1})] \hspace{-0.02in}- \epsilon'\{\partial U[x+R(\zeta_m^{n-1})] \odot \exp[Q(\xi_m^{n-1})] + T(\xi_m^{n-1})\}\right) \label{model def 1}\\
    \hspace{-0.1in}z^{n} &= \overline{m} \odot z^{n\prime} + m \odot \left(z^{n\prime} \hspace{-0.03in} \odot \exp[S(\xi_{\overline{m}}^{n\prime})] - \epsilon'\{\partial U\left[x+R(\zeta_{\overline{m}}^{n\prime})\right] \odot \exp[Q(\xi_{\overline{m}}^{n\prime})] + T(\xi_{\overline{m}}^{n\prime})\}\right)  \label{model def 2}
\end{align}
\normalsize
This update rule is similar to HMC in the sense that the final update to $x$ is generated by iteratively updating a variable with gradient evaluated at intermediate points. Here the intermediate points are learned with network $R$.

The log determinant of N steps of transformation is:
\begin{equation}
    \log \left|\frac{\partial z^{N}}{\partial z^0}\right| = \epsilon \ \mathbf{1} * \mathbf{1} + \sum_{n=1}^N \mathbf{1}*\left[\overline{m} \odot S(\xi_m^{n-1}\right)] + \mathbf{1}*\left[m \odot S(\xi_{\overline{m}}^{n'})\right]
\end{equation}
where $\mathbf{1}$ is the vector of $1$-entries with the same dimension as $z$.
The proposal entropy can be expressed as:
\begin{equation}
    \hspace{-0.1in}H\left(X'|X=x\right) \hspace{-0.03in}=\hspace{-0.03in} -\hspace{-0.03in}\int dx' q\left(x'|x\right) \log \left[q\left(x'|x\right)\right] \hspace{-0.03in}=  \hspace{-0.03in}- \hspace{-0.03in}\int\hspace{-0.03in} dz^0 p_{\mathcal{N}}\left(z^0\right) \left[\log \left(p_{\mathcal{N}}\left(z^0\right)\right) \hspace{-0.03in}-\hspace{-0.03in} \log \left|\frac{\partial z^{N}}{\partial z^0}\right|\right] \hspace{-0.05in}
\end{equation}
For each $x$, we aim to optimize $S(x) = \exp{\left[\beta H(X'|X=x)\right]} \times a(x)$, where $a(x)=\int A(x',x) q(x'|x) dx'$ is the average accept probability of the proposal distribution at $x$. Following \citep{dellaportas2019gradientbasedadaptiveMC}, we transform this objective into log space and use Jensen's inequality to obtain a lower bound: 
\begin{align*}
    \log{S(x)} &= \log{\int A(x',x)q(x'|x)dx'} + \beta H(X'|X=x) \\
               &\geq \int \log{[A(x'x)]}q(x'|x)dx' + \beta H(X'|X=x) = L(x)
\end{align*}
The distribution $q(x'|x)$ is reparameterizable, therefore the expectation over $q(x'|x)$ can be expressed as expectation over $p_{\mathcal{N}}(z_{0})$. Expanding the lower bound $L(x)$ and ignoring the entropy of the base distribution $p_{\mathcal{N}}(z_{0})$, we arrive at:
\begin{equation}
    L(x) = \int dz^0 p_{\mathcal{N}}(z^0)\left[\min{\left(0,\log{\frac{p(x')}{p(x)}} + \log{\frac{q(x|x')}{q(x'|x)}}\right)} - \beta \log \left|\frac{\partial z^{N}}{\partial z^0}\right|\right]
\end{equation}
During training we maximize $L(x)$ with $x$ sampled from the target distribution $p(x)$ if it's available, or with $x$ obtained from the bootstrapping process \citep{song2017A-NICE_MC} which maintains a buffer of samples and updates them continuously. Typically, only one sample of $z^0$ is used for each $x$.

A curious feature of our model is that during training one has to back-propagate over the gradient of target distribution multiple times to optimize $R$. In \citep{dellaportas2019gradientbasedadaptiveMC}, the authors avoid multiple back-propagation by stopping the derivative calculation at the density gradient term. In our experiment, we find it is necessary for good performance. 

The entropy-based exploration objective contains a parameter $\beta$ that controls the balance between acceptance rate and proposal entropy. We use a simple adaptive scheme to adjust $\beta$ to maintain a constant accept rate close to a target accept rate. The target accept rate is chosen empirically. As expected, we find the accept rate target needs to be lower for more complicated distributions.

\section{Related works: HMC and samplers it inspired}
\label{sec: related works}
Here we review HMC as well as other neural network MCMC samplers.
Methods we compare ours to in the Results are marked with {\bf bold font}.

First we revisit the {\bf HMC} sampler and show how it can be formulated as an M-H sampler. Basic HMC involves a Gaussian auxiliary variable $v$ of the same dimension as $x$, which plays the role of the momentum in Physics. HMC sampling consists of two steps: 1. The momentum is sampled from $\mathcal{N}(v; 0,\mathbf{I})$. 2. The Hamiltonian dynamics is simulated for a certain duration with initial condition $x$ and $v$, typically by running a few steps of the leapfrog integrator. Then, a M-H accept-reject process with accept probability $A(x',v',x,v) = \min{\left(1,\frac{p(x',v')q(x,v|x',v')}{p(x,v)q(x',v'|x,v)}\right)} = \min{\left(1,\frac{p(x')p_{\mathcal{N}}(v')}{p(x)p_{\mathcal{N}}(v)}\right)}$ is performed to correct for imprecision in the integration process. 
Since the Hamiltonian transition is \emph{volume-preserving} over $(x,v)$, we have $\frac{q(x,v|x',v')}{q(x',v'|x,v)} = 1$. Both HMC steps leave the joint distribution $p(x,v)$ invariant, and therefore HMC samples from the correct distribution $p(x)$ after marginalizing over $v$. To express basic HMC as a standard M-H scheme, step 1 and 2 can be aggregated into a single proposal distribution on $x$ to yield the standard M-H form with $q(x'|x) = p_{\mathcal{N}}(v)$ and $q(x|x')=p_{\mathcal{N}}(v')$. Note, although the probability $q(x'|x)$ can be calculated after the Hamiltonian dynamics is simulated, this term is intractable in general.  It is difficult to solve for the $v$ at $x$ to make the transition to $x'$ using the Hamiltonian dynamics. This issue is absent in RWM and MALA, where $q(x'|x)$ is tractable for any $x$ and $x'$.

{\bf A-NICE-MC} \citep{song2017A-NICE_MC}, which was generalized in \citep{spanbauer2020involutiveNN}, used 
the same accept probability as HMC, but replaced the Hamiltonian dynamics by a flexible volume-preserving flow \citep{dinh2014NICE}. A-NICE-MC matches samples from $q(x'|x)$ directly to samples from $p(x)$, using adversarial loss. This permits training the sampler on empirical distributions, i.e., in cases where only samples but not the density function is available. The problem with this method is that samples from the resulting sampler can be highly correlated because the adversarial objective only optimizes for the quality of the proposed sample. If the sampler produces a high quality sample $x$, the learning objective does not encourage the next sample $x'$ to be substantially different from $x$. The authors used a pairwise discriminator that empirically mitigated this issue but the benefit in exploration speed is limited. 

Another related sampling approach is {\bf Neural Transport MCMC} \citep{marzouk2016intrototransportMCMC,hoffman2019neutra,nijkamp2020flowbackboneEBM}
, which fits a distribution defined by a flow model $p_{g}(x)$ to the target distribution using $\mathbf{KL}[p_{g}(x)||p(x)]$. Sampling is then performed with HMC in the latent space of the flow model. 
Due to the invariance of the KL-divergence with respect to a change of variables, the ``transported distribution'' in $z$ space $p_{g^{-1}}(z)$ will be fitted to resemble the Gaussian prior $p_{\mathcal{N}}(z)$. 
Samples of $x$ can then be obtained by passing $z$ through the transport map. Neural transport MCMC improves sampling efficiency compared to sampling in the original space because a distribution closer to a Gaussian is easier to sample. However, the sampling cost is not a monotonic function of the KL-divergence used to optimize the transport map.  \citep{langmore2019conditionnumberHMC}. 

Another line of work connects the MCMC method to Variational Inference \citep{salimans2015VIandMCMCbridgegap,zhang2018ergodicMPF}. 
Simply put, they improve the variational approximation by running several steps of MCMC transitions initialized from a variational distribution. The MCMC steps are optimized by minimizing the KL-divergence between the resulting distribution and the true posterior. This amounts to optimizing a ``burn in'' process in MCMC. In our setup however, the exact sampling is guaranteed by the M-H process, thus the KL divergence loss is no longer applicable. Like in variational inference, the {\bf Normalizing flow Langevin MC} (NFLMC) \citep{gu2019NFLMC} also used a KL divergence loss. Strictly speaking, this model is a normalizing flow but not a MCMC method. We compare our method to it, because the model architecture, like ours, uses the gradient of the target distribution.

Another related technique is \citep{neklyudov2018MHviewofGANandVI}, where the authors trained an independent M-H sampler by minimizing $\mathbf{KL}\left[p(x)q(x'|x)||p(x')q(x|x')\right]$. This objective can be viewed as a lower bound of the M-H accept rate. 
However, as discussed in \citep{dellaportas2019gradientbasedadaptiveMC}, this type of objective is not applicable for samplers that condition on the previous state.

All the mentioned techniques have in common that their objective does not encourage exploration speed. In contrast, {\bf L2HMC} \citep{levy2017L2HMC,thinMetFlow} does encourage fast exploration of the state space by employing a variant of the expected square jump objective \citep{pasarica2010L2jump}: $L(x)= \int dx'q(x'|x)A(x',x)||x'-x||^2$. This objective provides a learning signal even when $x$ is drawn from the exact target distribution $p(x)$. L2HMC generalized the Hamiltonian dynamics with a flexible non-volume-preserving transformation \citep{dinh2016RNVP}. 
The architecture of L2HMC is very flexible and uses gradient of target distribution. However, the L2 expected jump objective in L2HMC improves exploration speed only in well-structured distributions (see Figure \ref{fig:Fig1 illustration of L2 and entropy loss}).

The shortcomings of the discussed methods led us to consider the use of an entropy-based objective. In principle, the proposal entropy objective could be optimized for the L2HMC sampler with variational inference \citep{poole2019variationalboundforMI,song2019MIestimators}, but our preliminary experiments using this idea were not promising. Therefore, we take inspiration from the Normalizing Flow model and investigated tractable ways to optimize the proposal entropy directly.

\section{Experimental result}
\subsection{synthetic dataset and bayesian logistic regression}
First we demonstrate that our technique accelerates sampling of the funnel distribution, a particularly pathological example from \citep{neal2003slicefunnel}. We then compare our model with A-NICE-MC \citep{song2017A-NICE_MC}, L2HMC \citep{levy2017L2HMC}, Normalizing flow Langevin MC (NFLMC) \citep{gu2019NFLMC} as well as NeuTra \citep{hoffman2019neutra} on several other synthetic datasets and a Bayesian logistic regression task. For all experiments, we report Effective Sample Size \citep{hoffman2014NUTS} per M-H step (ESS/MH) and/or ESS per target density gradient evaluation (ESS/grad). All results are given in \emph{minimum} ESS over all dimensions unless otherwise noted. We do not compare exact computation time or ESS/second, as commonly done in statistics literature, due to difficulties with reproducing previous models and with complications cased by using parallel computing hardwares (GPUs).


Here a brief description of the datasets used in our experiments:

{\bf Ill Conditioned Gaussian}: 50d ill-conditioned Gaussian task described in \citep{levy2017L2HMC}, a Gaussian with diagonal covariance matrix with log-linearly distributed entries between $[10^{-2}, 10^2]$.

{\bf Strongly correlated Gaussian}: 2d Gaussian with variance $[10^2,10^{-1}]$ rotated by $\frac{\pi}{4}$, same as in \citep{levy2017L2HMC}.

{\bf Funnel distribution}: The density function is $p_{funnel}(x)=\mathcal{N}(x_{0};0,\sigma^2) \mathcal{N}(x_{1:n};0,\mathbf{I} \exp{(-2x_{0})})$. This is a challenging distribution because the spatial scale of $x_{1:n}$ varies drastically depending on the value of $x_{0}$.  This geometry causes problems to adaptation algorithms that rely on a spatial scale. An important detail is that earlier work, such as \citep{betancourt2013softabs} used $\sigma=3$, while some recent works used $\sigma=1$. We run experiments with $\sigma=1$ for comparison with recent techniques and also demonstrate our method on a $20$ dimensional funnel distribution with $\sigma=3$. We denote the two variants by Funnel-1 versus Funnel-3.

{\bf Bayesian Logistic regression}: We follow the setup in \citep{hoffman2014NUTS} and used German, Heart and Australian datasets from the UCI data registry. 

\begin{figure}
\centering
\begin{floatrow}
\ffigbox[\FBwidth]
{
\caption{Comparison with HMC on the 20d Funnel-3 distribution. a) Chain and samples of $x_{0}$ (from neck to base direction) for HMC. b) Same as a) but for our learned sampler. Note, samples in a) look significantly more correlated than those in b), although they are plotted over a longer time scale.}
\label{fig: funnel-3 demo}
}
{
    \includegraphics[scale=0.35]{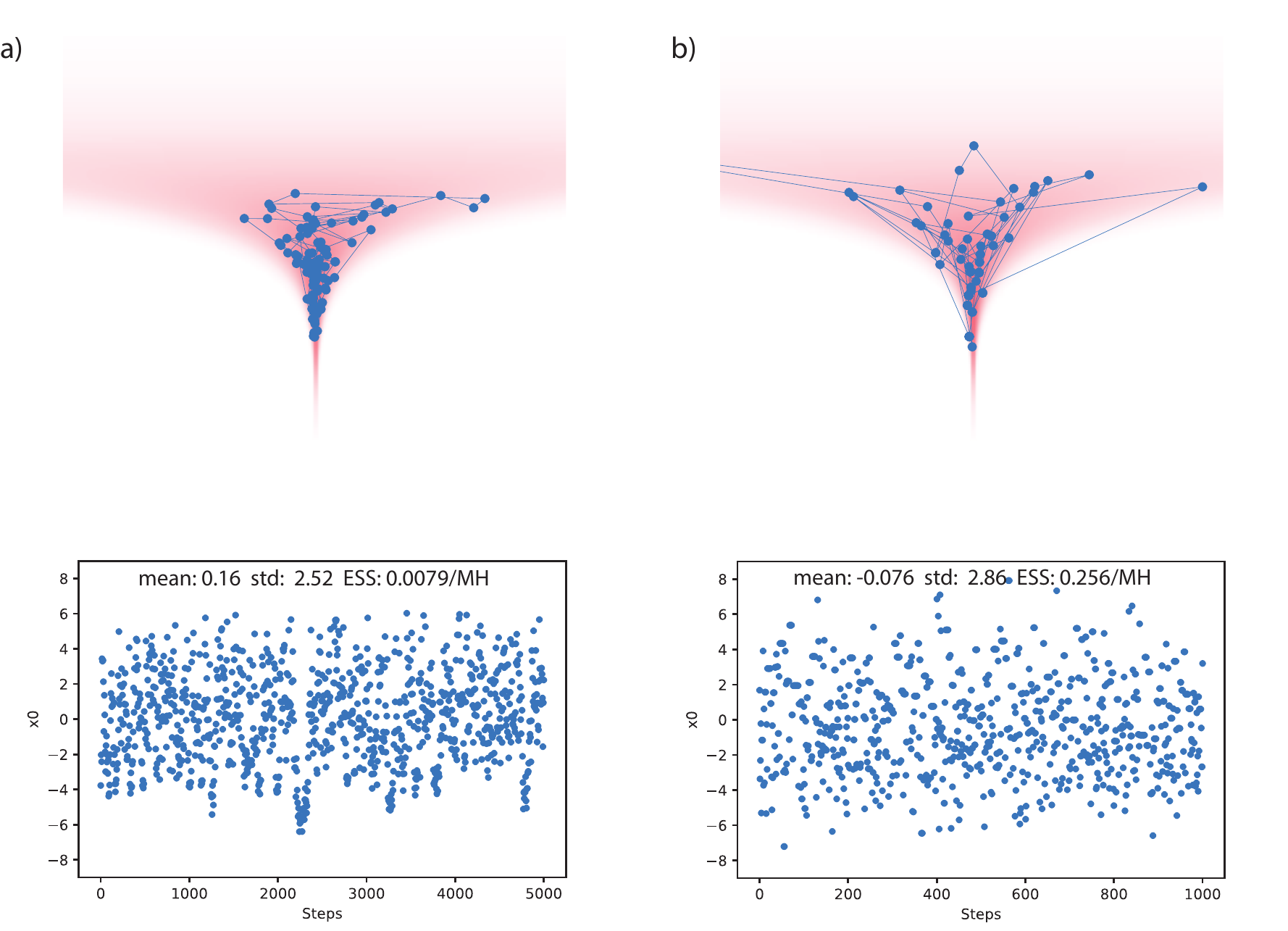}
}

\floatbox[\capbot]{table}[\Xhsize]{
\tiny
\begin{tabular}{c c c}
\multicolumn{1}{c}{\bf Dataset (measure)}  &\multicolumn{1}{c}{\bf L2HMC}  &\multicolumn{1}{c}{\bf Ours}
\\ \hline \\
50d ICG (ESS/MH)         &0.783  &{\bf 0.86} \\
2d SCG  (ESS/MH)           &0.497  &{\bf 0.89} \\
50d ICG (ESS/grad)       &$7.83 \times 10^{-2}$ &{$\bf 2.15 \times 10^{-1}$} \\
2d SCG (ESS/grad)       & $2.32 \times 10^{-2}$ &{$\bf 2.2 \times 10^{-1}$} \\
\hline \\
\multicolumn{1}{c}{\bf Dataset (measure)}  &\multicolumn{1}{c}{\bf Neutra}  &\multicolumn{1}{c}{\bf Ours}
\\ \hline \\
Funnel-1 $x_0$ (ESS/grad)       &$8.9 \times 10^{-3}$ &{$\bf 3.7 \times 10^{-2}$} \\
Funnel-1 $x_{1\cdots 99} $(ESS/grad)       & $4.9 \times 10^{-2}$ &{$\bf 7.2 \times 10^{-2}$} \\
\hline \\
\end{tabular}

\begin{tabular}{c c c c}
\multicolumn{1}{c}{\bf Dataset (measure)}  &\multicolumn{1}{c}{\bf A-NICE-MC} &\multicolumn{1}{c}{\bf NFLMC} &\multicolumn{1}{c}{\bf Ours}
\\ \hline \\
German (ESS/5k)       &926.49 & 1176.8 & {\bf 3150} \\
Australian (ESS/5k)       & 1015.75 & 1586.4  & {\bf 2950} \\
Heart (ESS/5k)      &1251.16  & 2000 & {\bf 3600}\\ 
\end{tabular}
}
{
\caption{Performance Comparisons. SCG: strongly correlated Gaussian, ICG: Ill-conditioned Gaussian. German, Autralian, Heart: Datasets for Bayesian Logistic regression. ESS: Effective Sample Size (a correlation measure)}
\label{table: performances}
}
\end{floatrow}
\vspace{-0.2in}
\end{figure}

In Figure \ref{fig: funnel-3 demo}, we compare our method with HMC on the 20d Funnel-3 distribution. As discussed in \cite{betancourt2013softabs}, the stepsize of HMC needs to be manually tuned down to allow traveling into the neck of the funnel, otherwise the sampling process will be biased. We thus tune the stepsize of HMC to be the largest that still allows traveling into the neck. Each HMC proposal is set to use the same number of gradient steps as each proposal of the trained sampler. As can be seen, the samples proposed by our method travel significantly further than the HMC samples. Our method achieves $0.256$ (ESS/MH), compared to $0.0079$ (ESS/MH) with HMC. 

As a demonstration we provide a visualization of the resulting chain of samples in Figure \ref{fig: funnel-3 demo} and the learned proposal distributions in Appendix \ref{fig: funnel-3 proposals}. The energy value for the neck of the funnel can be very different than for the base, which makes it hard for methods such as HMC to mix between them \citep{betancourt2013softabs}. In contrast, our model can produce very asymmetric $q(x'|x)$ and $q(x|x')$, making mixing between different energy levels possible.

Performances on other synthetic datasets and the Bayesian Logistic Regression are shown in Table \ref{table: performances}. In all these datasets our method outperformed previous neural network based MCMC approaches by a significant margin. Our model used various parameter settings, as detailed in Appendix \ref{sec: experimental details}. Results of other models are adopted or converted from numbers reported in the original papers. The Appendix provides further experimental results, ablation studies, visualizations and details on the implementation of the model.

\subsection{Training a convergent deep energy-based model}
A very challenging application of the MCMC method is training a deep energy-based model of images. In previous works, unadjusted Langevin dynamics was used to sample from the model. In addition to introducing bias, the step size had to be chosen carefully to achieve optimal performance \citep{nijkamp2019anatomyMCMCEBM,du2019implicit}. Here we apply our trainable sampler to train the EBM, which removes the need to choose the step size. The only tunable parameter is a target accept rate. We demonstrate stable training of a convergent EBM with significantly less sampling steps and gradient evaluations than previously reported. We also show that the learned sampler achieves better proposal entropy during training compared to the MALA algorithm. 

Similar to \citep{nijkamp2019anatomyMCMCEBM}, we use the Oxford flowers dataset of 8189 28$*$28 colored images. We dequantize the images to 5bits by adding uniform noise and use logit transform \citep{dinh2016RNVP}. Sampling is performed in the logit space. During training, we use Persistent Contrastive Divergence (PCD) \citep{tieleman2008PCD} with replay buffer size of 10000. 
We alternate between training the sampler and updating samples for the EBM training. Each EBM training step uses 40 sampling steps, with a target accept rate of $0.6$. Each EBM update use only 80 gradient evaluations, or equivalent to roughly 280 Langevin steps when sampler cost is taken into account. For comparison, the previous model used 500 Langevin steps. The savings in computation time will be even more pronounced in the training of more expressive EBMs, where gradient evaluation dominates computation cost.

\begin{figure}[th!]
    \centering
    \includegraphics[scale=0.85]{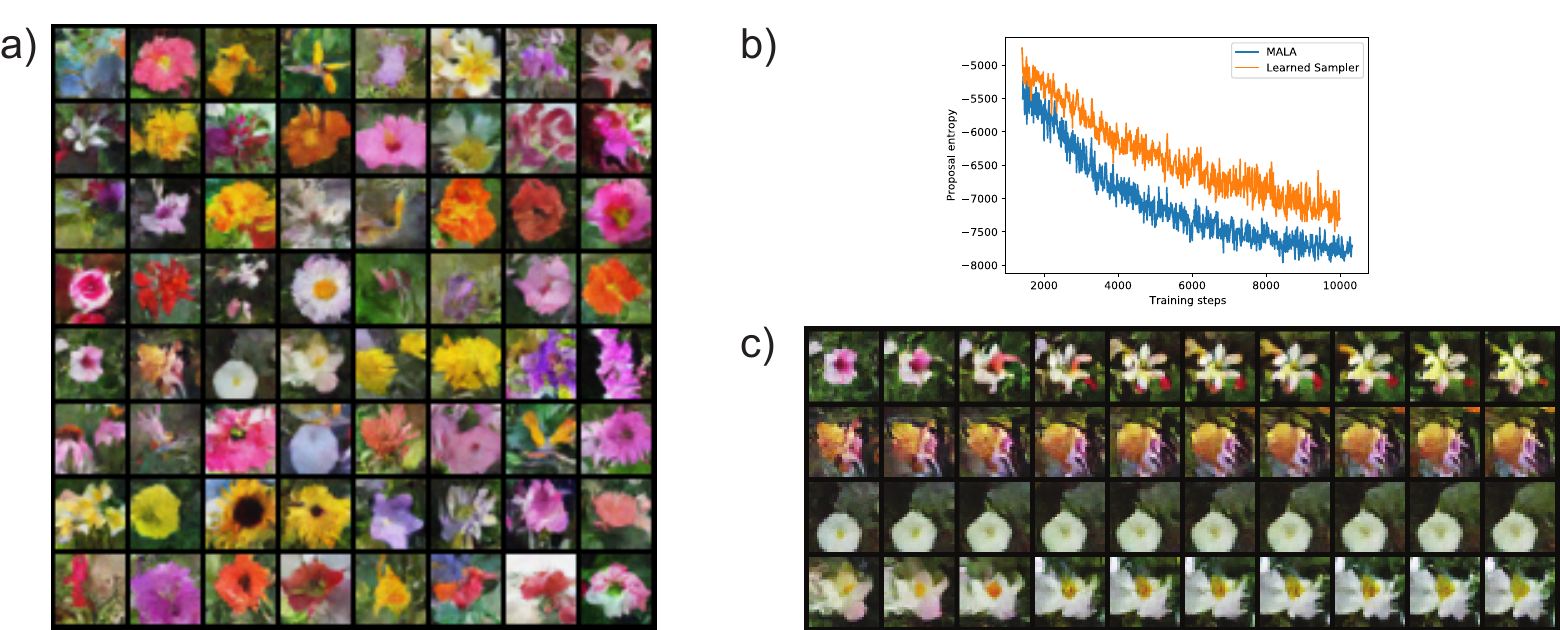}
    \caption{Training of convergent EBM with pixel space sampling. a) Samples from replay buffer after training. b) Proposal entropy of trained sampler vs MALA early during training, learned sampler is significantly higher. c) Samples from 100k sampling steps by the learned sampler, initialized at samples from replay buffer. Large transitions like the one in the first row is rare, here its selected for display.}
    \label{fig:Fig3 EBM result}
\end{figure}

Figure \ref{fig:Fig3 EBM result} depicts samples from the trained EBM replay buffer, as well as samples from a 100k step sampling process --for demonstrating stability of the attractor basins. We also compare the proposal entropy of the learned sampler during training with that of an adaptive MALA algorithm with the same accept rate target.

\section{Discussion}
In this paper we propose a gradient based neural network MCMC sampler with tractable proposal probability. The training is based on the entropy-based exploration speed objective. Thanks to an objective that explicitly encouraging exploration, our method achieves better performance than previous neural network based MCMC samplers on a variety of tasks. Compared to the RMHMC methods, our model provides a more scalable alternative for mitigating unfavorable geometry in the target distribution.  

There are many potential applications of our method beyond what was demonstrated in this paper. For example, training latent-variable models \citep{hoffman2017MCMClatentGaussianmodel}, latent sampling in GANs \citep{che2020GANEBMlatentsampling}, and others applications outside machine learning, such as molecular dynamics simulation \citep{noe2019boltzmanngenerator}. In the future, architectural improvement would also be interesting, use of auto-regressive architecture or different masking strategy may improve the expressiveness of our model. It will also be interesting to combine our technique with neural transport MCMC.

\subsubsection*{Acknowledgments}
We thank Hongye Hu for helpful suggestions on the manuscript. This work was funded by NSF award 1718991, NIH grant R01-EB026955 and by an INRC research grant from Intel Corporation.

\bibliography{iclr2021_conference}
\bibliographystyle{iclr2021_conference}

\newpage
\appendix
\counterwithin{figure}{section}
\counterwithin{table}{section}

\section{Appendix}
\subsection{Ablation study: effect of gradient information}
\label{sec: ablation}
In the main text we propose to use gradient information in sampling, inspired by the success of gradient-based sampling techniques. Here we present an ablation study that shows the impact of gradient information and of some architectural simplifications. Specifically we compare the full model to two variants: 1. the original model with data distribution gradient set to $0$, this is equivalent to generating proposals with a Real-NVP flow model conditioned on $x$. 2. A Langevin dynamics with noise generated by method 1. (See below)

More specifically, variant 2 consist of the following proposal process $x' = x + \epsilon z - \frac{\epsilon^2}{2} \partial_{x}U(x) \odot \exp{[S(x)]} + T(x)$, where $z=f(z_{0};x)$ is modeled by the original architecture with the gradient turned off, and $S$, $T$ are neural networks. This model amounts to Langevin dynamics with a flexible noise and an element-wise affine transformed gradient. It is not flexible enough to express the full covariance Langevin dynamics \cite{dellaportas2019gradientbasedadaptiveMC}, but we found its sufficient for energy-based model training, where it saves significant computation time because it only uses $2$ gradient evaluation per step instead of $4N$ as in the original model.

We train on the 100d Funnel-1 distribution, in each model the learning rate is tuned to the maximum stable value, with the same learning rate schedule. As can be seen from the training curve and visualizations, the model that does not use gradient information learns more slowly and has difficulty with mixing between energy levels, resulting in proposal distributions with poor coverage.

\begin{figure*}
    \label{fig: Grad compare}
    \centering
    \includegraphics[scale=0.82]{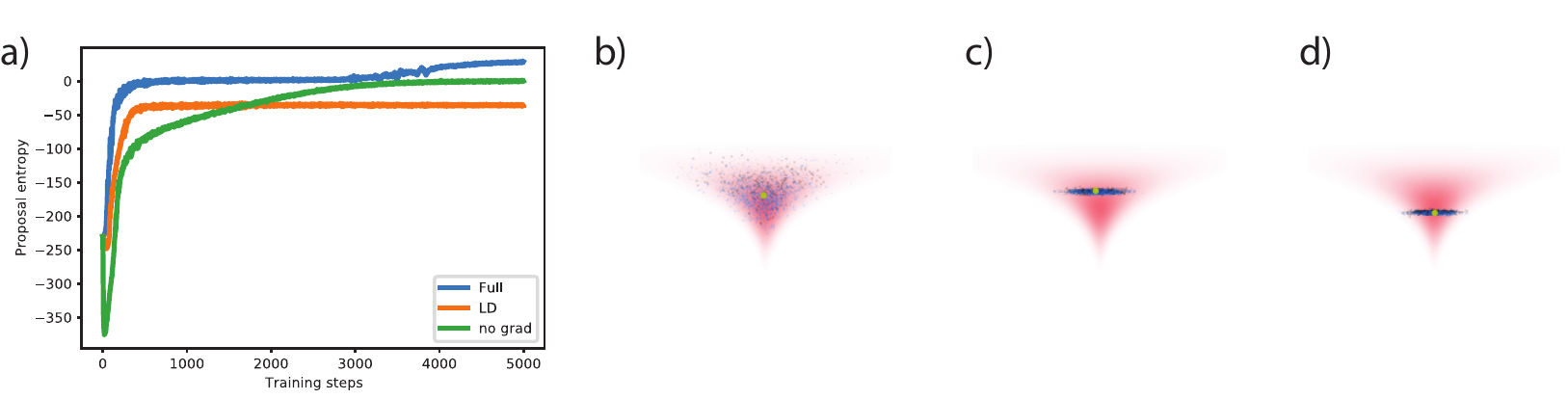}
    \caption{Illustrating the role of gradient information. a) Comparison of proposal entropy (up to a common constant) during training. Full: full model using gradient, LD: modification 2. discussed in the text, no grad: modification 1 in the text. b), c), d): Example proposal distribution of Full, LD and no grad model.}
\end{figure*}

\subsection{Experimental details}
\label{sec: experimental details}
Code for our model, as well as a small demo experiment, can be found under this link [\href{https://github.com/NEM-MC/Entropy-NeuralMC}{link}].

{\bf Architecture and training details} We use a single network for $S$, $Q$ and $T$. The weights of both, input and output layer, are indexed with step number as a means to condition on the step number. The two substeps of $z$ update need not to be indexed as they use disjoint sets of input and output units indicated by the masks. The weights of all other layers are shared across different steps. We use a separate network with the same architecture and size for $R$ and we condition on step number in the same way. Weights of all output layers of all networks are initialized at $0$. We use random masks that are drawn independently for each step.

When training the sampler, we use gradient clipping with L2 norm of $10$. Additionally, we use a cosine annealing learning rate schedule. For training the Bayesian logistic regression task, we use a sample buffer of size equal to the batch size. For ESS evaluation we sample for 2000 steps and calculate the correlation function for samples from the later 1000 steps. One exception is the HMC 20d Funnel-3 result, where we sampled for 50000 steps. For the ESS/5k result used in Bayesian logistic regression task, we simply multiply the obtained ESS/MH value by 5000. 

Other hyperparameters used in each experiments are listed in Table \ref{tab: hyperparas}. All models are trained with batch size of $8192$, for $5000$ steps, except for 20d Funnel-3, where the batch size is $1024$, and is trained for $20000$ steps. The momentum parameters in the Adam optimizer are always set to $(0.9,0.999)$. The parameter $\epsilon$ is chosen to be $0.01$ for EBM training and $0.1$ for all other experiments.

For deep EBM training we use variant 2, described in section \ref{sec: ablation} because it uses less gradient computation. We use a small 4-layer convnet with 64 filters for $S$, $Q$ and $T$. In the invertible network we use a fixed checkerboard mask. For EBM we used a 6 layer convnet with 64 filters, both networks use ELU activation function. In the EBM experiment we use 4 steps of $z$ updates. We use a replay buffer with 10000 samples, in each training step we first take a batch of 64 samples from the replay buffer, train the sampler for 5 consecutive steps before putting the samples back. If the average accept rate of the batches is higher than target accept rate $-0.1$, another 128 samples are draw from the replay buffer and are updated 40 times using the sampler. These samples are then used as negative samples in the Maximum Likelihood training of the EBM. The EBM uses the Adam optimizer with a learning rate of $10^{-4}$ and Adam momentum parameters of $(0.5,0.9)$. The sampler uses the same learning parameter, and has an accept rate target of 0.6. The EBM was trained for a total of 100000 steps, where the learning rate is decreased by a factor of $0.2$ at step $80000$ and $90000$. The proposal entropy and L2 expected jump of MALA in Figure \ref{fig:Fig3 EBM result} and Figure \ref{fig: appendix fig EBM} during training is obtained by loading an early checkpoint from the training process using the learned sampler, as MALA is unable to stably initialize the training. 

\begin{table}
    \centering
    \begin{tabular}{|c|c|c|c|c|c|}
         &  NN width & Steps & Accept rate target & Learning rate & Min learning rate \\
    \hline
    50d ICG         & 256 & 1 & 0.9 & $10^{-3}$ & $10^{-5}$  \\
    2d SCG          &  32  & 1 & 0.9 & $10^{-3}$ & $10^{-5}$  \\
    100d Funnel-1   & 512 & 3 & 0.7 & $10^{-3}$ & $10^{-5}$  \\
    20d Funnel-3    & 1024 & 4 & 0.6 & $5\times10^{-4}$ & $10^{-7}$  \\
    German          & 128 & 1 & 0.7 & $10^{-3}$ & $10^{-5}$  \\
    Australian      & 128 & 1 & 0.8 & $10^{-3}$ & $10^{-5}$  \\
    Heart           & 128 & 1 & 0.9 & $10^{-3}$ & $10^{-5}$  \\
    \end{tabular}
    \caption{Table for hyperparamters used in synthetic datasets and Bayesian Logistic regression. NN width: width of MLP network. Steps: Steps of updates in invertible model $f$. Min learning rate: terminal learning rate of cosine annealing schedule.}
    \label{tab: hyperparas}
\end{table}

{\bf Other datasets} Recent work on neural network based samplers used datasets other than those reported in the Results. We experimented with applying our model on some of those datasets. In particular, we attempted applying our method to the rotated 100d ill-conditioned Gaussian task in \citep{hoffman2019neutra}, without getting satisfactory result. Our model is able to learn the 2d SCG tasks which shows that it's able to learn a non-diagonal covariance, but in this task it learns very slowly and does not achieve good performance. We believe this is because coupling-based architectures do not have the right inductive bias to efficiently learn the strongly non-diagonal covariance structure, perhaps the autoregressive architecture used in \cite{hoffman2019neutra} would be more appropriate.

\citep{levy2017L2HMC} also presented the rough well distribution. We tried implementing it as described in the paper, but we found that a well-tuned HMC can easily sample from this distribution. Therefore we did not proceed to train our sampler on it. We should also add a comment regarding the 2d SCG task in \citep{levy2017L2HMC}. In the paper the authors stated that the covariance is $[10^2, 10^{-2}]$, but in the provided code it is $[10^2,10^{-1}]$, so we use latter in our experiment. 

Some other neural network MCMC studies consider the mixture of Gaussian distribution. Our model optimizes a local exploration speed objective starting from small initialization. It is therefore inherently local and not able to explore modes far away and separated by high energy barriers. The temperature annealing trick in the L2HMC paper \citep{levy2017L2HMC} does result in a sampler that can partially mix between the modes in a 2d mixture of Gaussian distribution. However, this approach cannot be expected to scale up to higher dimensions and more complicated datasets, therefore we didn't pursue it. It is our opinion the multi-modality of target distribution is a separate problem that probably should be solved by techniques such as independent M-H sampler \citep{neklyudov2018MHviewofGANandVI}, not by our current model. 

\subsection{Additional experimental results}
\label{sec: Additional results}

\begin{figure*}
    \centering
    \includegraphics[scale=0.82]{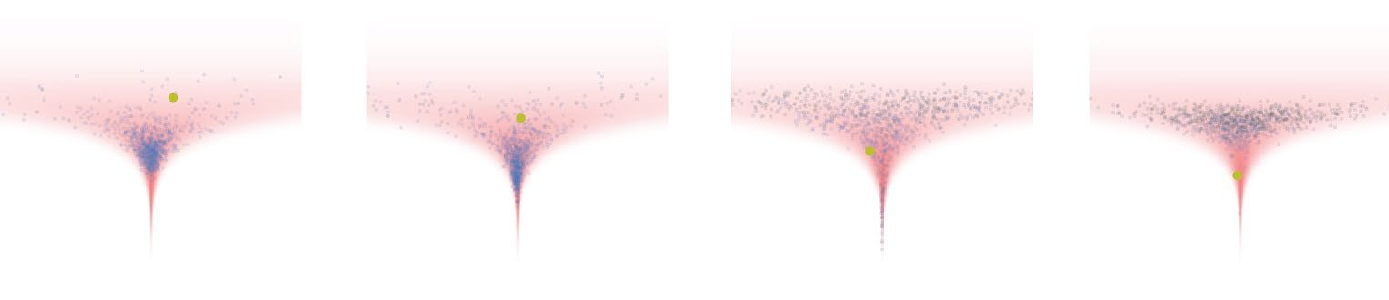}
    \caption{Some visualizations of proposal distributions learned on the Funnel-3 distribution. Yellow dot: $x$, Blue dots: accepted $x'$, Black dots: rejected $x'$. The sampler has an accept rate of ~0.6. Although not perfectly covering the target distribution, the proposed samples travel far from the previous sample and in a manner that complies to the geometry of the target distribution}
    \label{fig: funnel-3 proposals}
\end{figure*}

{\bf Visualization of Proposal distributions on Funnel-3 distribution}
We show the visualization of proposal distributions learned on the 20d Funnel-3 distribution in Figure \ref{fig: funnel-3 proposals}. This demonstrates the ability of the sampler to adapt to the geometry of the target distribution. Further, it shows that the sampler can generate proposal points across regions of very different probability density, since the neck and base of the funnel have very different densities but proposals can travel between them easily.

Our sampler can achieve significant speed up in terms of ESS/MH compared to HMC sampler, improvements comparable to Riemann Manifold HMC. However, the 100d funnel used in the manifold HMC paper still appears to be too difficult to our method, thus we used the easier 20d variant. 

\begin{figure*}[b!]
    \centering
    \includegraphics[scale=0.82]{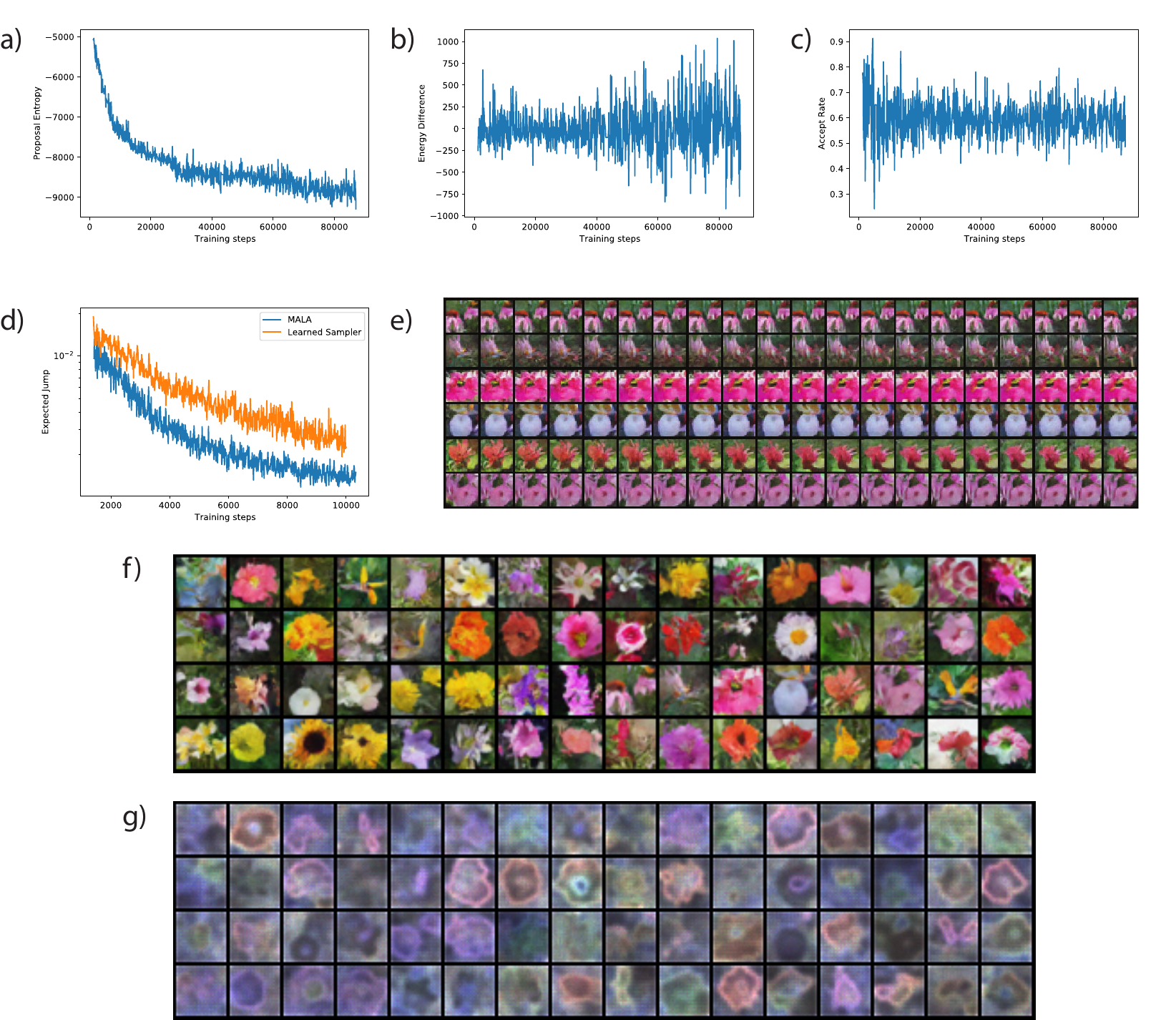}
    \caption{Further results for Deep EBM. a), b) and c): Proposal Entropy, Energy difference and accept rate curves during training. The sampler and EBM remain stable with a mixture of positive and negative energy difference. d) Comparison of L2 expected jump of MALA and learned sampler, plotted in log scale. It has almost the exact same shape as the proposal entropy plot in the main text. e) More samples from sampling process of 100k steps of learned sampler. f) g) Samples from the replay buffer and the corresponding visualization of the pixel-wise variance of displacement vector $z$ evaluated at the samples. Images in f) and g) are  arranged in the same order. Image-like structures that depends on the sample of origin are clearly visible in g). A MALA sampler would give uniform variance.}
    \label{fig: appendix fig EBM}
\end{figure*}

{\bf Further EBM results and discussion} 
Figure \ref{fig: appendix fig EBM} displays further results from the deep EBM training. Panels a) through c) demonstrate the stability of training, d) depicts the comparison of the L2 expected jump between learned sampler and MALA, f) and g) provide a sanity check. In g) of Figure \ref{fig: appendix fig EBM}, the pixel-wise standard deviation of variable $z=f(z_{0};x)$ (note it does not use gradient here as we use variant 2 in Appendix \ref{sec: ablation}) is displayed after normalizing it into image range. One can clearly see image-like structures similar to the sample of which the proposal is generated from. A MALA sampler would produce uniform images in this test, as $z$ is just a Gaussian in MALA. This shows that the learned sampler is utilizing the structures of the sample to generate better proposals. 

Our sampler provides more efficient exploration of the energy function \emph{locally}, as measured by the proposal entropy. However, as reported previously \cite{nijkamp2019anatomyMCMCEBM}, the learned energy landscape is highly multimodel with high energy barriers in between. Although a small level of mixing is visible in some sampling examples. Our sampler, being a local algorithm, cannot explore different modes efficiently, so we have the same non-mixing issue as previously reported. Additionally, as we do not use noise initialization, our model does not provide a meaningful gradient to re-sample from noise after the model has converged. This is quite different from what was reported in \citep{yu2020EBMwithfdivergence,grathwohl2019yourclassifierEBM}. The combination of none-mixing and inability of sampling from noise brings the problem of not being able to obtain new samples of the model distribution after the EBM is trained, replay buffer is all we have to work with. Resolving this difficulty is left for future study.

\end{document}